\documentclass[sigconf]{acmart}
\AtBeginDocument{%
  }
\copyrightyear{2023}
\acmYear{2023}
\setcopyright{rightsretained}
\acmConference[KDD '23]{Proceedings of the 29th ACM SIGKDD Conference on Knowledge Discovery and Data Mining}{August 6--10, 2023}{Long Beach, CA, USA}
\acmBooktitle{Proceedings of the 29th ACM SIGKDD Conference on Knowledge Discovery and Data Mining (KDD '23), August 6--10, 2023, Long Beach, CA, USA}
\acmDOI{10.1145/3580305.3599494}
\acmISBN{979-8-4007-0103-0/23/08}

\makeatletter
\gdef\@copyrightpermission{
  \begin{minipage}{0.3\columnwidth}
    \href{https://creativecommons.org/licenses/by-nd/4.0/}{\includegraphics[width=0.90\textwidth]{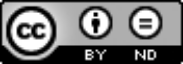}}
  \end{minipage}\hfill
  \begin{minipage}{0.7\columnwidth}
    \href{https://creativecommons.org/licenses/by-nd/4.0/}{This work is licensed under a Creative Commons Attribution-NoDerivs International 4.0 License.}
  \end{minipage}
  \vspace{5pt}
}
\makeatother

\usepackage{multirow}
\usepackage{subfigure}
\usepackage{algorithm}
\usepackage{algorithmicx}
\usepackage{algpseudocode}
\usepackage{color}
\usepackage{bm}
\begin{document}

\title{Rumor Detection with Diverse Counterfactual Evidence}


\author{Kaiwei Zhang}
\affiliation{%
  \institution{Institute of Information Engineering, Chinese Academy of Sciences}
  \institution{School of Cyber Security, University of Chinese Academy of Sciences}
  \city{Beijing}
  \country{China}
}
\email{zhangkaiwei@iie.ac.cn}
\orcid{0009-0006-8910-6283}

\author{Junchi Yu}

\affiliation{%
  \institution{MAIS\&CRIPAC, Institute of Automation, Chinese Academy of Sciences}
  \institution{University of Chinese Academy of Sciences}
  \city{Beijing}
  \country{China}
}
\email{yujunchi2019@ia.ac.cn}
\orcid{0000-0003-4118-3248}

\author{Haichao Shi}
\affiliation{%
  \institution{Institute of Information Engineering, Chinese Academy of Sciences}
  \city{Beijing}
  \country{China}
}
\email{shihaichao@iie.ac.cn}
\orcid{0000-0001-8846-1853}

\author{Jian Liang}
\affiliation{%
  \institution{MAIS\&CRIPAC, Institute of Automation, Chinese Academy of Sciences}
  \city{Beijing}
  \country{China}
}
\email{liangjian92@gmail.com}
\orcid{0000-0003-3890-1894}

\author{Xiao-Yu Zhang}
\affiliation{%
  \institution{Institute of Information Engineering, Chinese Academy of Sciences}
  \city{Beijing}
  \country{China}
}
\email{zhangxiaoyu@iie.ac.cn}
\orcid{0000-0003-1630-6058}
\authornote{Corresponding author.}

\renewcommand{\shortauthors}{Kaiwei Zhang, Junchi Yu, Haichao Shi, Jian Liang, \& Xiao-Yu Zhang}

\begin{abstract}
The growth in social media has exacerbated the threat of fake news to individuals and communities. This draws increasing attention to developing efficient and timely rumor detection methods. The prevailing approaches resort to graph neural networks (GNNs) to exploit the post-propagation patterns of the rumor-spreading process. 
However, these methods lack inherent interpretation of rumor detection due to the black-box nature of GNNs. Moreover, these methods suffer from less robust results as they employ all the propagation patterns for rumor detection.
In this paper, we address the above issues with the proposed \underline{D}iverse \underline{C}ounterfactual \underline{E}vidence framework for \underline{R}umor \underline{D}etection (DCE-RD).
Our intuition is to exploit the diverse counterfactual evidence of an event graph to serve as multi-view interpretations, which are further aggregated for robust rumor detection results. 
Specifically, our method first designs a subgraph generation strategy to efficiently generate different subgraphs of the event graph. We constrain the removal of these subgraphs to cause the change in rumor detection results. Thus, these subgraphs naturally serve as counterfactual evidence for rumor detection. To achieve multi-view interpretation, we design a diversity loss inspired by Determinantal Point Processes (DPP) to encourage diversity among the counterfactual evidence.
A GNN-based rumor detection model further aggregates the diverse counterfactual evidence discovered by the proposed DCE-RD to achieve interpretable and robust rumor detection results.
Extensive experiments on two real-world datasets show the superior performance of our method. Our code is available at  \href{https://github.com/Vicinity111/DCE-RD}{https://github.com/Vicinity111/DCE-RD}.

\end{abstract}




\maketitle

\section{Introduction}
The popularization of the Internet has dramatically changed the way in which people access news. The Internet has also become a huge melting pot of information. A large amount of unreliable information can quickly spread among people, which is prone to social chaos and disordering people's normal life. Rumor detection has gradually become a challenging task, especially in the face of public emergencies, widespread rumors can be extremely threatening and even cause death \cite{roozenbeek2019fake}. Meanwhile, due to the deliberate fabrication of rumor content and the complex propagation mode of rumors \cite{cheng2021causal}, it is difficult for people to intercept rumors.

Rumor detection is a subtask of text classification in natural language processing, which aims to identify the news texts without any structural rules and discriminate the rumors. However, only utilizing the textual content and ignoring structural information in post-propagation networks lead to the limitation of rumor detection models. Experiments conducted by \cite{kwon2017rumor, wu2015false, chen2016extreme, yang2018early,li-etal-2019-rumor-detection,huang2021mixgcf}show that models employing propagation networks perform better than those using textual-content-based features. Existing deep learning methods are proposed to model the propagation network by means of neural networks such as Graph Neural Networks (GNNs) \cite{bian2020, ma2017detect, rumor_yuan_2019, Min2022, nguyen2022, tian-etal-2022-duck}. These works have applied the GNN model for rumor detection to obtain the structures of post propagation networks. However, these methods have overlooked the \textbf{interpretation} of models, making the model performance improvement with limited space due to the black-box nature of GNNs. Unlike large Language models (LLMs), which have large model capacity and are trained with thousands of training data collected from various resources, most rumor detection models have less model capacity than LLMs and are trained with limited training data. Thus, the models are sensitive to the noise data and spurious patterns due to low model capacity and limited training data. This leads to degraded results and less robust performances. Besides, the studies of human cognition have also found that misinformation continues to spread even when it is discovered as people are more willing to believe debunked lies \cite{lewandowsky2012misinformation}. Therefore, it is important to provide great insights for understanding the spread of rumors \cite{roozenbeek2019fake}. Few works have developed the inherent interpretation of rumor detection models. 

If we assume that evidence is something that shows something else exists or is true, counterfactual explanations \cite{moraffah2020} are in the form of “If X had not occurred, Y would not have occurred” \cite{molnar2020interpretable}. \textbf{Counterfactual evidence} is the principled way to answer questions such as "if a particular part of an input graph is removed whether the prediction of the GNN model would change" and thus is highly desirable for GNNs. In the context of GNNs, removing a small fraction of nodes of the input graph identified by explanations can significantly change the predictions made by the GNNs. Counterfactual evidence is often concise and understandable \cite{moraffah2020,chang2023knowledge} since they are matched with human intuitions of described causal situations \cite{molnar2020interpretable}. To make explanations more robust and trustworthy, the counterfactual evidence should be \textbf{diverse}. Previous works \cite{wachter2017counterfactual, mothilal2020explaining} have presented a compelling argument highlighting the significance of diverse counterfactuals in informing a non-expert audience about the decisions that have been made. They stated that relying on a single explanation is limiting, as it only demonstrates how a decision was based on specific data that was both accurate and unchangeable by the decision-maker before future decisions, even if there exist other data that could be modified to achieve more favorable outcomes. 
Existing rumor detection models focused on generating subgraphs that are relevant for a particular prediction to explain the predictions made by GNNs. Few previous studies utilize counterfactual evidence for rumor detection. Given a prediction, the counterfactual evidence assists people in understanding how the prediction can be changed to achieve an alternative outcome\cite{lucic2022cf}.

In all, there are two main challenges for rumor detection. 
\par
(1) Propagation networks often contain numerous nodes and edges that are irrelevant to the task, e.g., posts that do not discuss the authenticity of news events. When applying graph neural networks on such structures, task-irrelevant information will be mixed into the neighborhood of nodes, which is known to reduce the generalization ability of subsequent classifiers \cite{zheng2020robust}. Exploiting key subgraphs for rumor detection alleviates this issue and improves the interpretation and performance of graph-based rumor detection by removing task-irrelevant nodes and edges.
\par
(2) How to generate diverse counterfactual evidence for rumor detection is a novel issue that has not been researched before. As discussed in Section \ref{2}, most GNN explanation methods are not counterfactual. These methods mainly focus on identifying subgraphs that are highly correlated with prediction results, which can not help identify the change required such that the prediction is changed. 

Our work resolves these challenges by developing a diverse counterfactual evidence framework (DCE-RD) for rumor detection. To be specific, we first design a subgraph generation strategy to generate explanations on each event graph, which aims to remove label-irrelevant information. We then discover diverse counterfactual evidence in rumor propagation graph, i.e. counterfactual subgraphs by removing the set of nodes included in the explanation of each event graph. Thus, these subgraphs naturally serve as counterfactual evidence for rumor detection. To achieve multi-view interpretation, we formulate a diversity loss inspired by Determinantal Point Processes (DPP) to encourage diversity among the counterfactual evidence. And DCE-RD further aggregates the diverse counterfactual evidence to achieve interpretable and robust rumor detection results. 
In this manner, our model can achieve good performance and achieve interpretation by revealing important counterfactual evidence for rumor detection.

In summary, our contributions are as follows:
\begin{itemize}
\item A \underline{D}iverse \underline{C}ounterfactual \underline{E}vidence framework for \underline{R}umor \underline{D}etection (DCE-RD) is proposed for rumor detection. Specifically, a subgraph generation strategy based on Top-K Nodes Sampling is proposed to efficiently produce subgraphs of the event graph. Then, counterfactual evidence is discovered by constraining the removal of these subgraphs to cause the change in rumor detection results.
\item A diversity loss inspired by determinantal point processes (DPP) is proposed to ensure the model extracts diverse substructures of post propagation networks. By promoting diversity of the counterfactual evidence, our model can leverage different critical substructures of rumor propagation to achieve multi-view interpretation of graph-based rumor detection. DCE-RD further aggregates the diverse counterfactual evidence to achieve interpretable and robust rumor detection results.
\item Experimental results including comparison experiments,  robustness evaluation and early rumor detection on two real-world datasets demonstrate the effectiveness and robustness of DCE-RD. 
\end{itemize}

\section{related works}\label{2}

In recent years, there has been an increasing amount of literature on rumor detection. It can be divided into two categories according to the information used for rumor detection: \textbf{textual-content-based} and \textbf{propagation-network-based}. Textual-content-based methods, which utilize the source post and all user replies, explore the stance and opinion of users towards rumors. Deceptive events usually have a different content style from the truth, often using exaggerated emotional expressions to deceive readers.
Castillo et al. \cite{castillo2011information} used many text features in their model, such as the
fraction of tweets with and crafted features. These features
and other additional text features are also used in \cite{liu2015real, enayet-el-beltagy-2017-niletmrg, li-etal-2019-rumor-detection}. 
Chua and Banerjee \cite{chua2016linguistic} analyzed
six groups of features: comprehensibility,
sentiment, time-orientation, quantitative details,
writing style, and topic.
Ma et al. \cite{ma2015detect} considered the variation of these social context features during the message propagation over time. These approaches use statistical models or deep learning models of events and tweet content, sometimes added with detailed user profiles, to classify rumors as true or false.
However, only considering the textual content leads to the ignorance of propagation structure in social networks, which could limit the detection performance of models.
Propagation-network-based methods construct flexible networks to capture structural information about rumor propagation and utilize the path of re-shares and other propagation dynamics for rumor detection. 
There are many works exploiting various mechanisms to model the rumor propagation pattern and combining textual-content data with structural information. 
Gupta et al. \cite{gupta2012evaluating} constructed a multi-typed network consisting of events, tweets, and users. Jin et al. \cite{jin2014} proposed a hierarchical credibility network consisting of an event, sub-events and messages for revealing vital information for rumor detection, and they further developed credibility network with conflicting relations \cite{jin2016news}. 
Kernel methods are designed to evaluate the similarity between two events. Wu et al. \cite{wu2015false} proposed a random walk graph kernel to model the similarity of propagation trees. Ma et al. \cite{ma2017detect} proposed a Propagation Tree Kernel (PTK) to differentiate false and true rumors by evaluating the similarities between their propagation tree structures, as tree kernel is specifically well-suited for structured data. Rosenfeld et al. \cite{rosenfeld2020} presented an efficient implementation of the Weisfeiler-Lehman graph kernel and explored whether only focusing on topological information can be predictive of veracity.
Deep learning approaches including user-attention-based convolutional neural network model \cite{ma2018rumor}, RNN-based method \cite{liu2018early}, LSTM-based model \cite{zubiaga2018detection}, and GAN-style approach \cite{ma2019detect} achieve good predictive performance on rumor datasets.
With the significant advancement of graph neural networks (GNNs), most existing methods have attempted to integrate GNNs with rumor detection to explore the post propagation pattern such as \cite{bian2020, ma2017detect, rumor_yuan_2019, Min2022, nguyen2022, tian-etal-2022-duck}. However, these works overlooked the interpretation of the graph-based models. Our work focuses on improving the interpretation of graph-based model for rumor detection.

\textbf{Interpretation} in graph learning is in need as explaining predictions made by GNNs helps collect insights from graph-structured data \cite{wu2022survey,zhang2022trustworthy}. Recently, an increasing number of approaches have been proposed to provide interpretation for GNNs. 
Chang et al. \cite{chang2020invariant} proposed a game-theoretic invariant rationalization criterion to identify a small subset of input features as best explains or supports the prediction. 
Yu et. al. \cite{yu2020graph,yu2021recognizing} proposed a Graph Information Bottleneck (GIB) framework to recognize the maximally informative yet compressive subgraphs for graph interpretation.
To address the training inefficiency and instability of GIB, VGIB \cite{yu2022improving} and GSAT \cite{miao2022interpretable} employ the variational inference to leverage the reduction of stochasticity to select label-relevant subgraphs for graph interpretation. Chen et. al. \cite{chen2022learning} and Yu. et al. \cite{yu2023mind} further enhance the generalization performance of GNNs by improving graph interpretation.

Inspired by methods that achieve better performance by improving graph interpretation, most approaches for rumor detection focus on identifying subgraphs as explanations for prediction. Yang et al. \cite{yang2022reinforcement} proposed a reinforced subgraph generation method, and perform fine\-grained modeling on the generated subgraphs. 
Yet these methods are generally not \textbf{Counterfactual} in nature: Given a prediction, a small part of perturbation to the input graph could change the prediction \cite{lucic2022cf}. Counterfactual evidence is often concise and easy to understand \cite{moraffah2020}, as it aligns with human intuitions of causal situations. Mothilal et al. \cite{molnar2020interpretable} developed a novel method to provide post-hoc explanations of machine learning models utilizing counterfactual explanations. Our approach aims to extend an intriguing class of explanations through counterfactual evidence for graph-based models and applies to the task of rumor detection. And previous works \cite{wachter2017counterfactual, molnar2020interpretable} have presented the significance of diverse counterfactuals in informing a non-expert audience about the decisions that have been made. Relying on a single explanation is limiting as there may exist other data that could be changed to achieve a more favorable outcome.
To address the aforementioned issues, we comprehensively utilize both textual content and propagation structure for rumor detection. Additionally, our work focuses on discovering diverse counterfactual evidence by recognizing multiple counterfactual subgraphs in the rumor propagation graph. By promoting diversity of the counterfactual evidence, our model can leverage different important substructures of rumor propagation to achieve multi-view interpretation of graph-based rumor detection.

\section{method}

\begin{figure}[htbp]
	\centering
	\begin{minipage}[c]{0.6\columnwidth}
		\centering
		\includegraphics[width=\columnwidth]{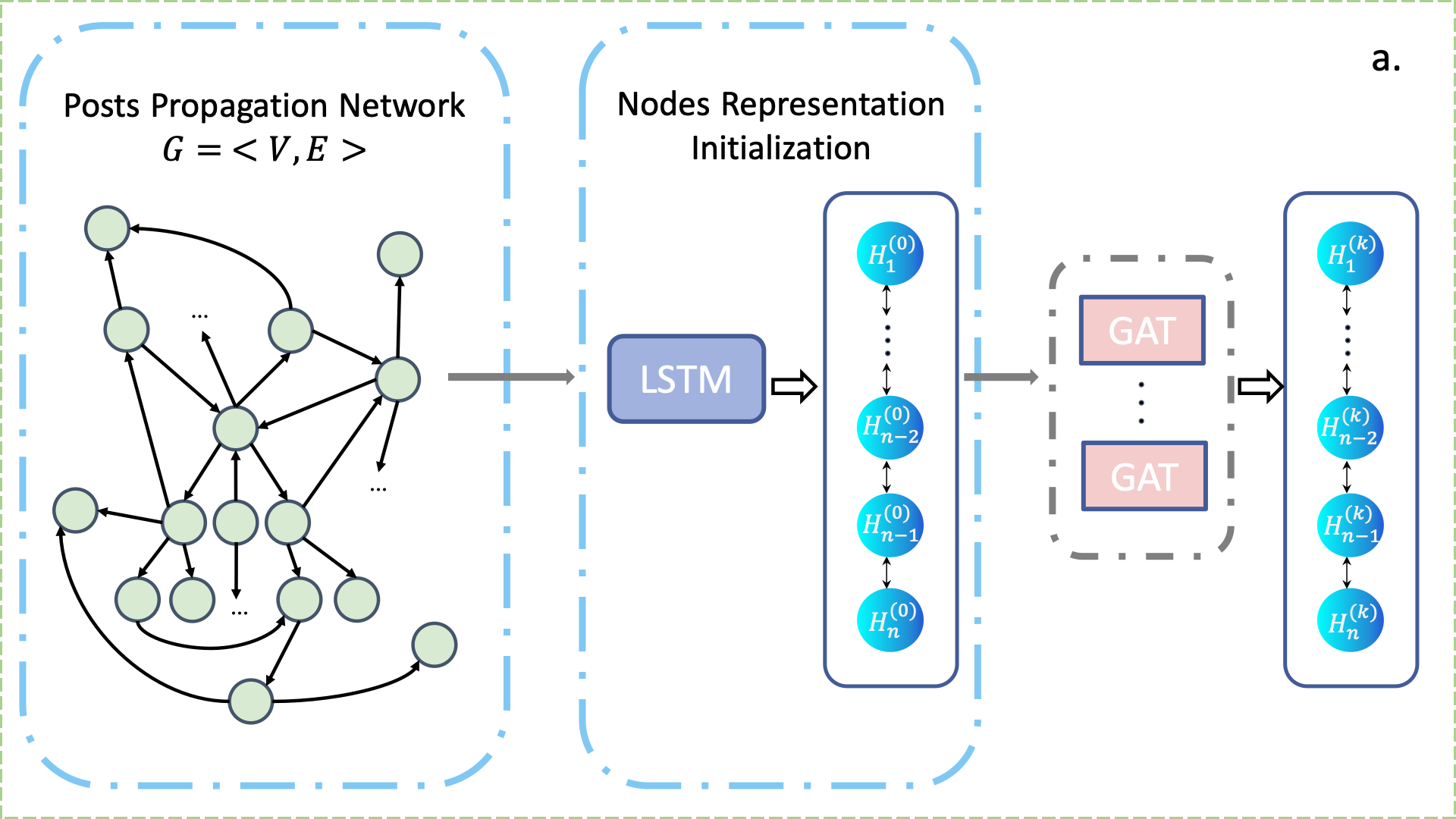}
            \vspace{-4mm}
	\end{minipage}
        \begin{minipage}[c]{0.3\columnwidth}
		\centering
		\includegraphics[width=\columnwidth]{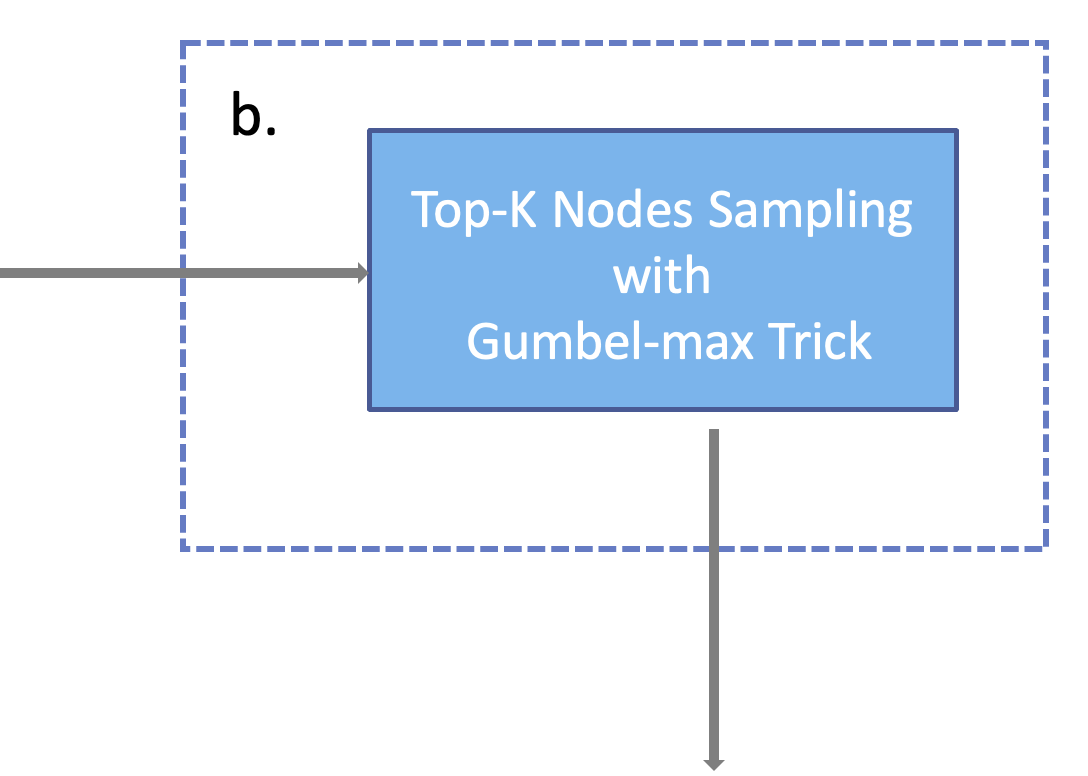}
            \vspace{-14mm}
	\end{minipage}

	\begin{minipage}[c]{0.45\columnwidth}
		\centering
		\includegraphics[width=\columnwidth]{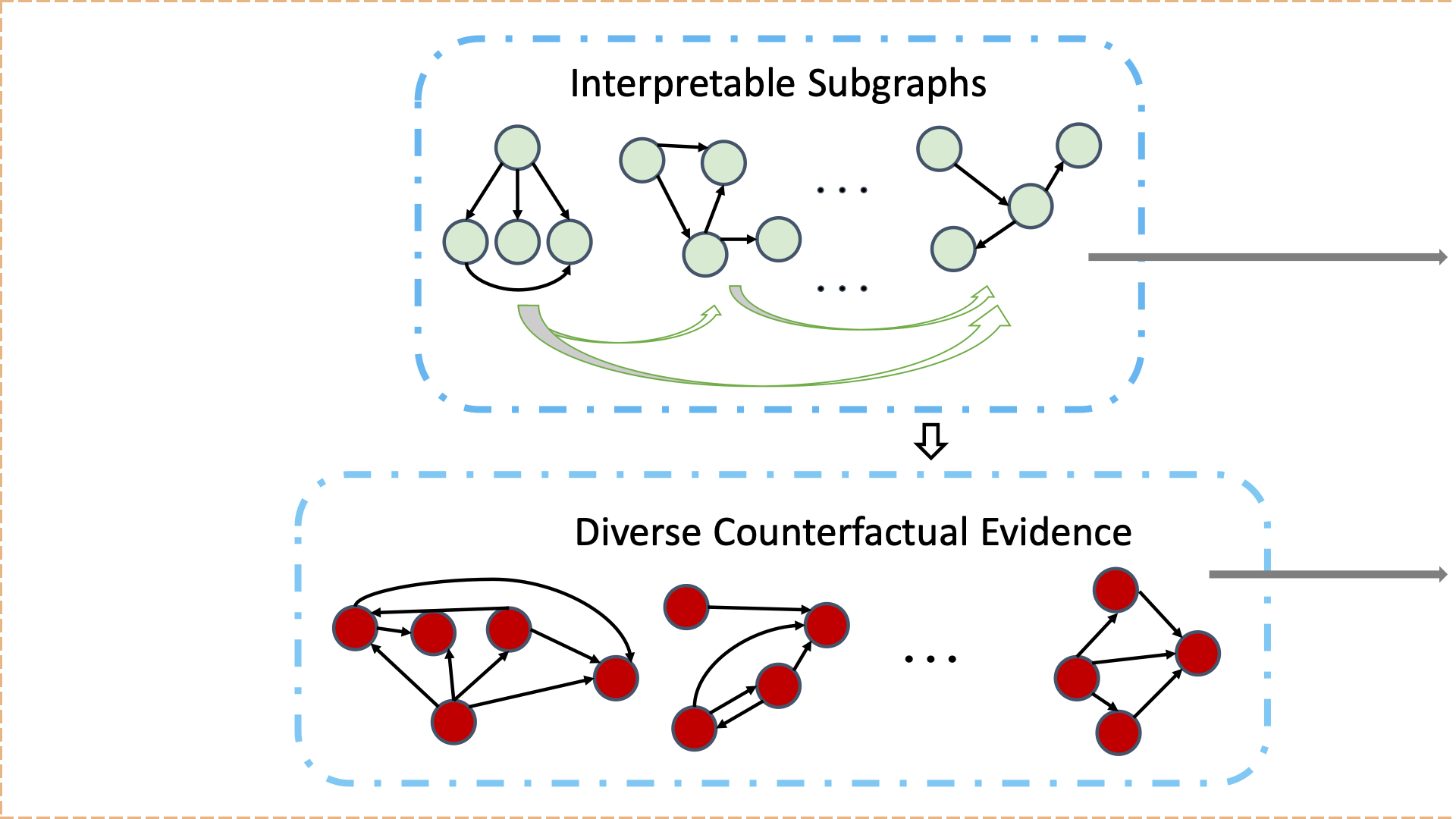}
	\end{minipage} 
	\begin{minipage}[c]{0.45\columnwidth}
		\centering
		\includegraphics[width=\columnwidth]{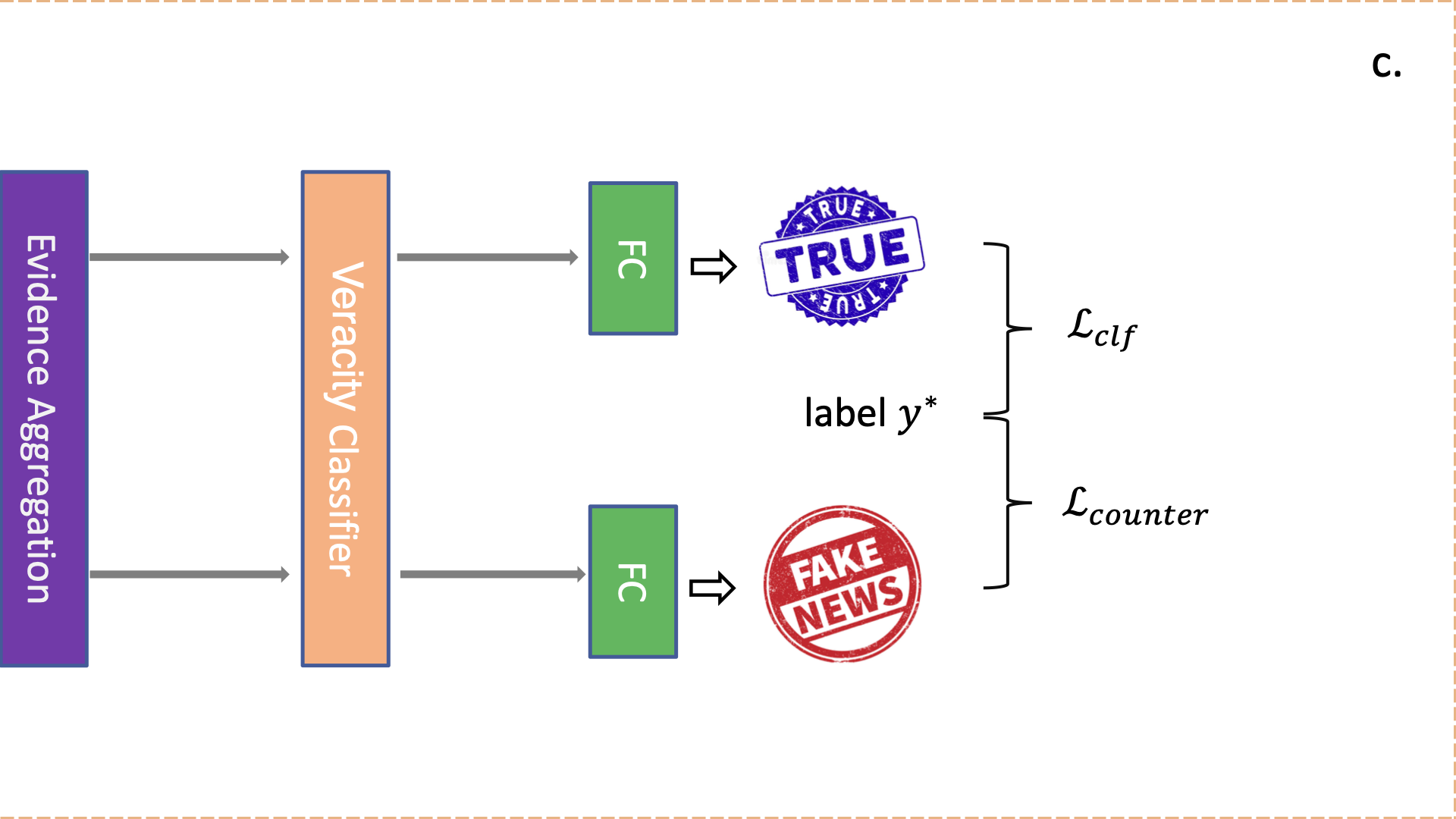}
	\end{minipage}
	\caption{Our proposed Diverse Counterfactual Evidence Framework for Rumor Detection (DCE-RD). (a). We first initialize the representation of each node in the event graph. GAT represents Graph Attention Network for post propagation network modeling. (b). Then we perform Top-K Nodes Sampling with Gumbel-max Trick on each event graph to produce subgraphs. (c). Afterward, we constrain the removal of these subgraphs from the event graph to generate counterfactual evidence, which causes the change in rumor detection results. Additionally, diverse generated counterfactual evidence is further aggregated to achieve multi-view interpretation and robustness. The corresponding representations are fed into a veracity classifier, respectively. }
	\label{fig:Arch}
\end{figure}

In this section, we introduce a \underline{D}iverse \underline{C}ounterfactual \underline{E}vidence framework for \underline{R}umor \underline{D}etection, named as DCE-RD. The core idea of DCE-RD is
to generate diverse subgraphs and counterfactual evidence based on Top-K node sampling with Gumbel-max Trick for rumor detection. Additionally, we formulate a novel loss function based on Determinantal Point Processes (DPP) as a form of robust optimization.
The architecture of our method is shown in Figure \ref{fig:Arch}. 

\subsection{Problem Formulation}

Let $R=\{R_1, R_2, \cdots, R_N\}$ be a rumor dataset, where $R_i$ is the $i-$th rumor event and $N$ is the total count of rumor events. We define 
$R_i=\{n_i, P_i, G_i, G_{sub}, G^{\perp}_{sub}\}$, where $P_i=\{p^i_1, p^i_2, \cdots, p^i_{n_i}\}$ refers to the set of posts related to the event and $n_i$ is the total count of posts, $G_i$ represents the posts propagation structure network, $G_{sub}$ refers to the set of diverse subgraphs. Then removing the set of nodes identified by the $G_{sub}$ from the $G_i$ to generate the corresponding set of counterfactual evidence $G^{\perp}_{sub}$. 
Specifically, $G_i = <V_i, E_i>$ is defined as a graph, where $V_i= P_i$ represents the posts contained in the event and $E_i=\{e^i_{st} | s,t=0,1,\cdots,n_i-1\}$ refers to the edges from the source post $p_i^{s}$ to the responsive post $p_i^{t}$, i.e., retweeted post or reply post \cite{wu2015false, Ma2016}. For example, if $p^i_3$ retweeted $p^i_4$, there will be an directed edge $p^i_4 \to p^i_3$, i.e., $e^i_{43}$. If $p^i_2$ replied the $p^i_0$, there will be an directed edge $p^i_0 \to p^i_2$, i.e., $e^i_{02}$. For each rumor event $R_i$, there is a label $y_i$ related to its veracity, i.e., $y_i=1$ represents True Rumor and $y_i=0$ represents False Rumor \cite{ma2017detect, zubiaga2018detection}. The model of rumor detection is to calculate the final probability prediction $\hat{y}$.

Thus, the goal of rumor detection is achieved by minimizing the cross-entropy loss:
\begin{equation}
    \mathcal{L} = - y \log(\hat{y})  + (1-y)\log(1- \hat{y}), 
\end{equation}

\textbf{Definition 1.} \textit{Counterfactual Evidence.} Given a rumor propagation network $G$ of a rumor event, the set of sampled subgraphs $G_{sub}$, and its label $y$. The Counterfactual evidence $G^{\perp}_{sub}$ seeks for the maximum drop in the confidence of the original prediction. 

The set of sampled subgraphs $G_{sub} = \{s_1, s_2, \cdots, s_m\}$ is formed through selecting the important nodes based on $Top-K$ Nodes Sampling with Gumbel-max Trick, each element in the set $s_j \in G_{sub}$ is a connected subgraph which serves as the reason for predicting the label $y$. Counterfactual evidence $G^{\perp}_{sub}$ is generated by removing the set of nodes in $G_{sub}$ from $G$ individually, which causes the maximum drop in the confidence of the original prediction. Similarly, counterfactual evidence serves as the reason for changing the label. We focus on optimizing the counterfactual property of evidence along with the association between the evidence and the prediction. Thus, both $G_{sub}$ and $G^{\perp}_{sub}$ are critical for rumor detection. The details are shown in Sec \ref{3.2}. 

\subsection{Counterfactual Evidence Generation}\label{3.2}

This section is the answer to the question \textit{how to generate interpretable subgraphs and counterfactual evidence for rumor detection}.

\subsubsection{Node Representation Initialization}

Textual content is a strong indicator used to spot potential deception. Rumors usually gain public attention through malicious guidance or exaggerated emotional expression \cite{Min2022}. Thus, the emotional tendency, bot-like flag or political stance of posts could also imply the veracity of posts. There are many deep learning networks to represent textual content in natural language processing, such as LSTM \cite{shi2015convolutional}, GRU \cite{cho2014learning}, Transformer \cite{vaswani2017attention}, Bert \cite{kenton2019bert} and GPT \cite{radford2018improving}. In our work, the representation of each node $v_i$ is initialized by using its textual embedding $h_i$ and we apply LSTM (Long-Short Term Memory) \cite{shi2015convolutional} as textual feature extractor to obtain joint features of text and social context, which shows the best performance in our experiment. We then obtain the event text embedding matrix $ \bm{H_R} =\left\{ h_1, \cdots, h_n\right\}$ for any event $R$, where $h_i$ is the hidden states at the last layer of LSTM for $i$-th post. 

\subsubsection{Top-K Nodes Sampling}

Inspired by Top-K sampling \cite{RSS2019}, we present Top-K Nodes Sampling to sample $K$ nodes from a graph of $n$ nodes. We consider finite $n$ and only produce samples after processing the entire graph. In sampling, each node $v_i$ is associated with a weight $w_i \ge 0$. In our experiment, we regard the attention score matrix as a weight matrix. Let $\bm{W} = \left[ w_1, \cdots, w_n \right]$ and $S = \sum_{i=1}^n w_i$. Let $\bm{Z^j} = \left[ z^j_1, \cdots, z^j_n \right]$ be a one-hot vector, i.e., if one nonzero element at index $j$ in a vector, $z^j_j=1$. Let $S_{ws} = \left[ \bm{Z^{i_1}}, \cdots, \bm{Z^{i_K}} \right]$, a $K$ length sequence of one-hot vectors, where $\bm{Z^{i_j}}$ represents selecting $v_{i_j}$ in the $j-$th sample. We wish to sample $S_{ws}$ from

\begin{equation}
    P(S_{ws} | \bm{W}) = \frac{w_{i_1}}{S}\frac{w_{i_2}}{S-w_{i_1}}\cdots\frac{w_{i_K}}{S-\sum_{j=1}^{K-1} w_{i_j}}, 
\end{equation}
which corresponds to sampling without replacement with
probabilities proportional to item weights. Each node $v_i$ is
given a random key $r_i = u_i^{1/w_i}$ where $u_i$ is drawn from a
uniform distribution between $\left[0, 1\right]$. Let the top $K$ keys over the $n$ nodes be $r_{i_1}, \cdots, r_{i_K}$ and $S_{ws} = \left[ \bm{Z^{i_1}}, \cdots, \bm{Z^{i_K}} \right]$ is associated with the top $K$
keys. The output of sampling is distributed according to $P(S_{ws} | \bm{W})$ \cite{efraimidis2006}. We define the function $Top$-$K(V, \bm{W}, K)$ as the Top-K Nodes Sampling. 

\textbf{Gumbel-max Trick}
The Gumbel-max trick \cite{yellott1977} generates random keys $\hat{r_i} = \log (w_i) + g_i$ by
perturbing logits with Gumbel noise $g_i \sim Gumbel(0, 1)$,
then taking $v_{i*}$ such that $i* = argmax_i \hat{r_i}$ as a sample.
These samples are distributed based on $p(v_i|\bm{W}) = \frac{w_i}{\bm{S}}$.
The Gumbel-softmax trick makes training with backpropagation possible \cite{maddison2016, jang2016} by reparameterizing the sample as a deterministic transformation of the parameters $\bm{W}$ and some independent noise $g_i$ and relaxing the deterministic transformation
(from max to softmax). The sampling details are shown in Algorithm \ref{Algorithm 1}.

\begin{algorithm}
	\renewcommand{\algorithmicrequire}{\textbf{Input: }}
	\renewcommand{\algorithmicensure}{\textbf{Output:}}
	\caption{Top-K Nodes Sampling, $Top$-$K(V, \bm{W}, K)$}
	\label{Algorithm 1}
	\begin{algorithmic}[1]
 
            \Require Nodes Set $V = \left\{ v_0, \cdots, v_n \right\}$, Weight matrix $\bm{W} = \left[ w_1, \cdots, w_n\right] $, Subset size $K$.

            \State Initialization: $\hat{r} \leftarrow []$

            \For {$i:=1$ to $n$}
                \State $g_i \sim Gumbel(0,1)$
                \State $r_i \leftarrow \log(w_i) + g_i$
                \State $\hat{r}.append(r_i)$
            \EndFor
                
            \State $c = \sum_{j=1}^K \bm{Z^{i_j}}$, where $i_j$ is the $j$-th top key in $\hat{r}$
            
		\Ensure  $K$-hot vector $c = \left[ c_1, \cdots, c_n \right]$, where $\sum_{i=1}^n c_i = K$.
	\end{algorithmic}  
\end{algorithm}

\subsubsection{Counterfactual Evidence Generation}

We propose a subgraph generation strategy based on Top-K nodes sampling to allow label-relevant information kept in subgraphs and further generate counterfactual evidence, which significantly improves interpretation of the model.

\textbf{Node-level attention} Given an event graph of posts, the adjacent matrix $A_R$ and the post feature matrix $H_R^0 = \left\{ h^{(0)}_1, \cdots, h^{(0)}_n \right\} = H_R$, our method computes the attention score of each node through combining its textual content and topological information. Specifically, the attention score between $i-$th node and $j-$th node can be formulated
as:
\begin{equation}
  e_{ij}^{(l)}= \vec{b}^TLeakyReLU(W_a^{(l)}\cdot\left[ h_i||h_j \right]), 
\end{equation}
\begin{equation}
    a_{ij}^{(l)}=Softmax(e_{ij}^{(l)})=\frac{exp(e_{ij}^{(l)})}{\sum_{k \in \mathcal{N}(i)} exp(e_{ik}^{(l)})}, 
\end{equation}
where $\vec{b} \in \mathbb{R}^d$ is a parameter vector, $W_a^{(l)} \in \mathbb{R}^{2d}$ a learnable parameter matrix to project nodes representation, $e_{ij}^{(l)}$ and $a_{ij}^{(l)}$ are unnormalized and normalized attention
between the adjacent nodes $v_i$ and $v_j$. After computing the attention scores for all neighbor nodes, the central node’s representation is
updated by aggregating features weighted by the attention scores:

\begin{equation}
    h^{(l+1)}_i = \sigma(\sum_{j \in \mathcal{N}(i)} a^{(l)}_{ij}W_u^{(l)}h_j^{(l)}), 
\end{equation}
where $\sigma$ is a nonlinear function and $W_u$ is a learnable update matrix parameter. Here, we adopt ReLU function as the activation function. Dropout \cite{srivastava2014dropout} is applied on
GraphConv Layers to avoid over-fitting. However, the update process can cause long-distance dependency. In our method, we add residual networks to update the equation:

\begin{equation}
     h^{(l+1)}_i = \sigma(\sum_{j \in \mathcal{N}(i)} a^{(l)}_{ij}W_u^{(l)}h_j^{(l)})+h_i^{(l)}, 
\end{equation}

Here, we adopt a modified Graph Attention Network (a.k.a GATv2) \cite{brody2021attentive} as our backbone to compute the representation of the graph, which fixes the static attention problem of the standard Graph Attention Network \cite{velivckovic2017graph} 
and show more robust performance. Then, we obtain $\tilde{H}_R = H^k_R=\left\{ h_0^{(k)}, \cdots, h_n^{(k)} \right\}= \left\{ \tilde{h_0}, \cdots, \tilde{h_n} \right\}$ after $k$ layers of GATv2 representation. 

\textbf{Subgraph Generation} After $m$ iterations of Top-K Nodes sampling, we obtain a set of subgraphs $G_{sub} = \{s_1, s_2, \cdots, s_m\}$, where $s_i$ is formed from $Top$-$K(V, \tilde{H}_R, K)$ in the $i-$th iteration. Analogously, the set of counterfactual evidence $G^{\perp}_{sub}$ is formed as:
\begin{equation}
    G^{\perp}_{sub} = \{G\setminus s_1, G\setminus s_2, \cdots, G\setminus s_m\}, 
\end{equation}
where $G\setminus s_i$ represents the complement subgraph of all nodes which are not in the subgraph $s_i$. Note that the source post is always included in the  $G\setminus s_i$. And we only consider connected subgraphs in our experiment. Then we could obtain the embedding of subgraph $s_i$, referred to $\tilde{H}_{s_i}$.

\subsection{Diversity of Evidence}\label{3.3}

Previous work \cite{wachter2017counterfactual} has presented a compelling argument highlighting the significance of diverse counterfactuals in informing a non-expert audience about the decisions that have been made. Relying on a single explanation is limiting, as it only demonstrates how a decision was based on specific data that was both accurate and unchangeable by the decision-maker before future decisions, even if there exist other data that could be modified to achieve a more favorable outcome. 
Our experimental results (Tables \ref{tab:McFake Comparison} and \ref{tab:MaWeibo Comparison}) also show that diverse counterfactual evidence is better than single one. In this section, the question \textit{how to ensure the diversity of multiple counterfactuals so that they provide multi-view interpretation} is to be answered.

\subsubsection{Determinantal Point Processes (DPP) }

For a given set $Z$, DPP is a probability measure defined over all subsets of that set. For example, a set containing $N$ elements $\boldsymbol{y} = \left\{ 1,2,\cdots,N \right\}$, there are $2^N$ possible sub-sets and DPP defines a probability measure $\mathcal{P}$ over subsets $Y \subseteq \boldsymbol{y}$.
DPP converts the complex probability calculation in the Discrete Point Processes problem into a simple determinant calculation, and calculates the probability of each subset through the determinant of the kernel matrix \cite{borodin2009determinantal, kulesza2012determinantal}.

For example, there is a real type, PSD (positive semi-definite) matrix $L$, for any subset $Y$ of the set $Z$, the probability of occurrence of $Y$ is proportional to the determinant of matrix $L_Y$: 

\begin{equation}
    \mathcal{P}(Y) \propto det(L_Y), 
\end{equation}
where, the number of rows and columns of the matrix $L$ are the number of elements of the set $Z$, the matrix $L$ can be indexed by the elements of $Z$, and the matrix $L_Y$ is the matrix obtained by indexing the matrix $L$ through the elements of the subset $Y$. The condition of PSD (positive semidefinite) guarantees that the determinants of all submatrices of the kernel matrix are non-negative \cite{chen2018fast, kulesza2012determinantal}.

Our diversity loss encourages the sampler to generate diverse subgraphs which keep the label-relevant information and critical substructure of the input graph. However, merely increasing the similarity between generated subgraphs and input graph will lead to counterfactuals that are too close in the representation field. Instead, we are interested in increasing diversity within generated subgraphs for generating diverse counterfactual evidence. Therefore, we build a DPP kernel for each subgraph during every iteration of the training process. To simplify learning a kernel, we match the eigenvalues and eigenvectors of a subgraph DPP kernel with the other one until the end of total enumeration. The eigenvalues and vectors capture the manifold structure of subgraphs, thus making optimization more feasible.

We use Determinantal Point Processes (DPP) kernel to evaluate the diversity between subgraphs. Based on DPP-inspired loss, we extend a method to define diversity loss for subgraphs:

\begin{equation}
    \mathcal{L}_{diversity} = \sum_k \left \| \lambda_i^k - \lambda_j^k \right\|_2 - \sum_k v_i^k{v_j^k}^T,
\end{equation}
where $\lambda_i^k$ and $\lambda_j^k$ are the $k-$th eigenvalues of $\tilde{H}_{s_i}$ and $\tilde{H}_{s_j}$ respectively. Similarly, $v_i^k$ and $v_j^k$ are the $k-$th eigenvectors of $\tilde{H}_{s_i}$ and $\tilde{H}_{s_j}$ respectively.

\begin{algorithm}
	\renewcommand{\algorithmicrequire}{\textbf{Input: }}
	\renewcommand{\algorithmicensure}{\textbf{Output:}}
	\caption{Evidence Aggregation and Prediction}
	\label{Algorithm 2}
	\begin{algorithmic}[1]
 
            \Require $G_{sub} = \{s_1, s_2, \cdots, s_m\}$, corresponding embedding representation $\tilde{H}_{s_1}, \cdots, \tilde{H}_{s_m}$.

            \State Initialization: $D[0:m,0:m] \leftarrow []$
            \For {$i:=0$ to $m$}
                \For  {$j:=0$ to $m$}
                    \If{$i \ne j$}
                    \State $D[i][j] \leftarrow \left \| \tilde{H}_{s_i} - \tilde{H}_{s_j} \right\|_2 $
                    \EndIf
                \EndFor
                \State $L_{s_i} \leftarrow \sum_j \log(D[i][j])$
                
            \EndFor
            \State $Ls \leftarrow concat(L_{s_0}, \cdots, L_{s_m})$ \Comment{concat by row}
            \State $W_{clf} \leftarrow Softmax(Ls)$
		\State Calculate prediction probabilities based on Equation~(\ref{6})
		
		\Ensure  $\mathbb{P}(y | G_{sub}, G^{\perp}_{sub} )$.
	\end{algorithmic}  
\end{algorithm}

\subsection{Rumor Detection with Aggregated Evidence}

Given diverse interpretable evidence, we aggregate evidence for rumor detection with the classification weight matrix calculated by the embedding of each evidence. Specifically, we weighted the influence of each evidence on the prediction. Thus the label for the event could be predicted using aggregated evidence. The details are shown in Algorithm \ref{Algorithm 2}.

\subsubsection{Evidence Aggregation} 

We propose a method for aggregating evidence, which also applies to counterfactuals aggregation. For any pair of subgraphs $(s_i, s_j), i\neq j$, the distance matrix of their embeddings is formed as:

\begin{equation}
    D_{ij} = \left \| \tilde{H}_{s_i} - \tilde{H}_{s_j} \right\|_2 ,
\end{equation}
where $\left \| \cdot \right\|_2$ refers to Euclidean norm (2-norm). And,

\begin{equation}
    Ls_i = \sum_{j \neq i } \log(D_{ij}),
\end{equation}
where $Ls_i$ is the sum distance matrix of subgraph $s_i$. And we define $Ls = \left\{ sum(Ls_0, 0), \cdots, sum(Ls_m, 0) \right\}^T$, $sum(Ls_i, 0)$ is the result of summing each column of the input tensor. Then, we calculate the classification weight matrix which represents the influence of each evidence on the prediction.

\begin{equation}
    W_{clf} = Softmax(Ls),
\end{equation}

\subsubsection{Classification} 

Our model evaluates the probability of each class and selects the label with the highest probability as the final prediction result. We define the classification probability matrix as follows:

\begin{equation}\label{6}
    \mathbb{P}(y | G_{sub}) =
    Softmax(W_s^T(ReLU(W_f(W_{clf}^T \tilde{H}_{sub})+b_f))+b_s),
\end{equation}
where $\tilde{H}_{sub} = Concat\left\{ \tilde{H}_{s_0}, \cdots, \tilde{H}_{s_m}\right\}$, and $W_f, b_f, W_s, b_s$ are learnable parameters during training. Here we adopt ReLU function as the activation function. Dropout \cite{srivastava2014dropout} is applied to avoid over-fitting.

\subsubsection{Model Optimization}

The total loss is divided into diversity loss, classifier loss, and counterfactual loss. Diversity loss has been introduced in Sec\ref{3.3}. Classifier loss and counterfactual loss are introduced in this section. 

\textbf{Classifier Loss} The aim of our sampler $f_\tau$ with parameter $\tau$ is  to extract predictive subgraphs. $f_\tau$ blocks the label-irrelevant information while allowing the label-relevant information kept in subgraphs to make predictions. We define classifier loss via cross-entropy loss:

\begin{equation}
    \mathcal{L}_{clf} = - y^* \log\mathbb{P}(y | G_{sub})  + (1-y^*)\log(1- \mathbb{P}(y | G_{sub}), 
\end{equation}
where $y^* \in \left\{ 0,1 \right\}$ is the ground-truth label of the event and $G_{sub} \sim f_\tau (G)$. 

\textbf{Counterfactual Loss} Similarly, we define counterfactual loss via cross-entropy loss:
\begin{equation}
    \mathcal{L}_{conter} = - \bar{y} \log\mathbb{P}(y | G^{\perp}_{sub})  + (1-\bar{y}) \log(1- \mathbb{P}(y | G^{\perp}_{sub} )), 
\end{equation}
where $\bar{y}$ is the opposite of ground-truth label of the event, which refers to the change of the prediction. 

Finally, we obtain the total loss:
\begin{equation}
    \mathcal{L} = \alpha \mathcal{L}_{diversity} + \beta \mathcal{L}_{clf} + \gamma \mathcal{L}_{counter}, 
\end{equation}
where $\alpha, \beta, \gamma$ are hyper-parameters. Generally, take $\alpha, \beta, \gamma \in \left[ 0,2 \right]$.

DCE-RD provides interpretable explanations for predictions by sampling nodes to form connected subgraphs in order to remove redundant information and retain information related to prediction in subgraphs. Additionally, DCE-RD focuses on
discovering diverse counterfactual evidence by recognizing multiple counterfactual subgraphs in the rumor propagation graph. During training, the proposed loss function promotes the diversity among counterfactual evidence and further improves the performance of classifier. Promoting diversity of the counterfactual evidence can leverage different important substructures of rumor propagation to
achieve multi-view interpretation of graph-based rumor detection. 

\section{experiments and results}

In this section, we first compare with several baseline models to evaluate the performance of our proposed method DCE-RD. Then, Gaussian noise is added to node feature to investigate the robustness of our method. Finally, we examine the capability of early rumor detection for both DCE-RD and the compared methods.

\begin{table}
  \caption{The statistics of the MC-Fake dataset.}
  \label{tab:McFake Statistics}
  \vskip 0mm
  \begin{center}
  \begin{tabular}{l|cc}
    \toprule
    Labels & True Rumors & False Rumors\\
    \midrule
    Events Count & 3381 & 4966\\
    \cline{1-3}
    Connections Count & 686,700 & 849,362\\
    \cline{1-3}
    Users Count & 849,305 & 1,165,510\\
    \cline{1-3}
    Tweet Count & 991,764 & 1,348,868\\
    \cline{1-3}
    Retweet Count & 442,863 & 580,810\\
    \cline{1-3}
    Reply Count & 292,272 & 317,655\\
    \cline{1-3}
    Avg. \# of Tweet Count / event & 293 & 272 \\
    \cline{1-3}
    Avg. \# of Retweet Count / event & 131 & 117\\
    \cline{1-3}
    Avg. \# of Reply Count / event & 86 & 64\\
  \bottomrule
\end{tabular}
\end{center}
\end{table}

\begin{table}
  \caption{The statistics of the Ma-Weibo dataset.}
  \label{tab:MaWeibo Statistics}
  \vskip 0mm
  \begin{center}
  \begin{tabular}{l|c}
    
    \toprule
    Events Count & 4664\\
    \cline{1-2}
    True Rumors & 2351\\
    \cline{1-2}
    False Rumors & 2313\\
    \cline{1-2}
    Connections Count & 3,805,656\\
    \cline{1-2}
    Users Count & 2,746,818\\
    \cline{1-2}
    Avg. \# of Posts Count / event &  816\\
    \cline{1-2}
    Max. \# of Posts Count / event & 59,318\\
    \cline{1-2}
    Min. \# of Posts Count / event & 10\\
  \bottomrule
\end{tabular}
\end{center}
\end{table}

\subsection{Datasets} 

Weibo and Twitter are the most popular social media in China and the U.S., respectively. We evaluate our proposed model on two real-world datasets: Weibo \cite{Ma2016} and MC-Fake \cite{Min2022}.
The MC-Fake dataset is a recently proposed dataset with more abundant news events (28334 events) compared to Twitter15 \cite{liu2015real, Ma2016} and Twitter16 \cite{ma2017detect}, including corresponding social contexts (tweets, retweets, replies, retweet\_relations, replying\_relations) and diverse topics (Politics, Entertainment, Health, Covid-19, Syria War) collected from Twitter via Twitter API \footnote{https://developer.twitter.com/en}. This dataset has more recent data with the emergence of topics like Covid-19. Thus, we employ MC-fake dataset for evaluation.
However, MC-Fake contains huge blank data, i.e., disconnected graph, thus we further refined it. To refine MC-Fake, we first extracted the data items relevant to our experiment (tweets, retweets, replies, retweet relations, and replying relations). Then we removed events with only a single post since they cannot be connected into graphs or sampled to form subgraphs and these single-post events were in the minority of the dataset. Instead, we utilized the majority of events collected from Twitter which have a significant amount of interaction. 
In these two datasets, nodes refer to posts, edges represent retweet or reply relationships between two nodes. Both Weibo and MC-Fake contain two binary labels: False Rumor (F) and True Rumor
(T). The label of each event in Weibo is
annotated according to Sina community management center,
which reports various misinformation \cite{Ma2016}.
And the label of each event in MC-Fake is
annotated according to the veracity tag of the article in rumor
debunking websites \footnote{e.g., snopes.com, emergent.info,
etc} \cite{ma2017detect}. The statistics of the MC-Fake and the Weibo are shown in Table \ref{tab:McFake Statistics} and Table \ref{tab:MaWeibo Statistics}, respectively.

\subsection{Baselines}  

We compare the proposed model DCE-RD with some state-of-the-art baselines, including:

\begin{itemize}
\item DTC \cite{castillo2011information}: A rumor detection
method using a Decision Tree classifier based on
various handcrafted features to obtain information credibility.

\item SVM-RBF \cite{yang2012automatic}: A SVM-based model with
Radial basis function (RBF) kernel, using handcrafted features based on the overall
statistics of the posts.

\item SVM-TS \cite{ma2015detect}: A linear SVM classifier that
leverages handcrafted features to construct time-series
model.

\item PPC\_RNN+CNN \cite{liu2018early}: A rumor detection model based on RNN and CNN, which learns the rumor
representations through the characteristics of users in the
rumor propagation path.

\item GLAN \cite{rumor_yuan_2019}: A global-local attention network for rumor detection, which jointly utilizes the local semantic and global structural information. To make a fair comparison, we only focus on the post propagation network yet user social network.

\item SureFact \cite{yang2022reinforcement}: A reinforced subgraph generation framework, which develops a hierarchical path-aware kernel graph attention network to perform fine-grained modeling on the generated subgraphs. 

\item PSIN \cite{Min2022}: A fake news detection
model, which adopts a divide-and-conquer strategy to model the post-post, user-user and post-user interactions in social context. We also focus on the post-post interactions so as to make a fair comparison.

\item UPFD-GCN \cite{kipf2016semi}: A scalable approach for semi-supervised learning on graph-structured data that is based on an efficient variant of convolutional neural networks which operates directly on graphs. 

\item GCNFN \cite{monti2019fake}: A novel automatic
fake news detection model based on geometric deep learning. The core algorithms are a generalization of classical convolutional neural networks to graphs, allowing the fusion of heterogeneous data. 

\item DCE-RD ($\alpha=0$): Our proposal DCE-RD without diversity loss.

\item DCE-RD ($m=1$): Our proposal DCE-RD with only one evidence for prediction.
\end{itemize}

\subsection{Experiment Settings}

We implement DTC and SVM-based models with scikit-learn \footnote{https://scikit-learn.org/stable/}. And
PPC\_RNN+CNN, GLAN, SureFact, PSIN, UPFD-GCN, GCNFN, and our method are implemented with Pytorch \footnote{https://pytorch.org/}. To ensure a stable and comprehensive evaluation, we adopt a five-fold cross-validation strategy. The dataset was split into 5 folds, we pick 4 folds for training, and 1 fold for testing.
We evaluate the Accuracy (Acc.), F1 measure
(F1), AUC \cite{fawcett2006introduction} over the two categories and Precision (Prec.), Recall (Rec.), F1 measure (F1) on each class. 

The parameters of our method are updated using mini-batch gradient descent, and we optimize the model by Adam algorithm \cite{kingma2014adam}. The batch size is $128$. The dimension of each node's hidden state is $64$. The number of neural network layers for each part is selected from $\left\{ 2, 3, 4 \right\}$. The rate of dropout is $0.5$. The learning rate is selected from $\left\{ 0.5, 0.05, 0.005, 0.0005 \right\}$. The process of training is iterated upon $100$ epochs. We select $K = \lceil \kappa \times n \rceil $ nodes from a graph to form a subgraph, $\kappa$ refers to the ratio of the number of nodes in a subgraph to the total number of nodes in the original graph. Typically, $\kappa$ is selected from $0.1$ to $0.9$. After $m$ iterations of the above process, $m$ subgraphs are generated to represent the diverse evidence. And $m$ is usually selected from $\left\{ 2, 3, 4, 5, 6 \right\}$. Hyper-parameters $\alpha, \beta, \gamma$ is selected from $0$ to $2$. Early stopping \cite{yao2007early} is applied when the test loss
stops decreasing by $15$ epochs.

\begin{table*}
  \caption{Comparison with baselines on the MC-Fake dataset. M1 represents non-graph-based methods, and M2 refers to graph-based methods. Best baseline results are
highlighted with underlined text. Best overall results are marked by *. Statistically significant improvement (t-test over 5 different dataset splits, p-value$<$ 0.05) is highlighted with bold text. We evaluate the Accuracy (Acc.), F1 measure (F1), and AUC over the two categories and Precision (Prec.), Recall (Rec.), and F1 measure (F1) on each class.}

  \label{tab:McFake Comparison}
  \begin{tabular}{c|cc|cccccccc|cc|cc}
    \toprule
    \multicolumn{1}{c|}{}&
    \multicolumn{2}{c|}{\multirow{2}{*}{Method}} &
    \multicolumn{2}{c|}{\multirow{2}{*}{ACC}} & \multicolumn{2}{c|}{\multirow{2}{*}{F1}} & \multicolumn{2}{c|}{\multirow{2}{*}{AUC}} & \multicolumn{2}{c|}{Prec} & \multicolumn{2}{c|}{Rec} & \multicolumn{2}{c}{F1}\\
    \cline{10-15}
    \multicolumn{1}{c|}{}&
    \multicolumn{2}{c|}{} & \multicolumn{2}{c|}{} & \multicolumn{2}{c|}{} & \multicolumn{2}{c|}{} & T & F & T & F & T & F\\
    \midrule
    \multicolumn{1}{c|}{\multirow{4}{*}{M1}}&
    \multicolumn{2}{c|}{DTC} & \multicolumn{2}{c|}{0.7552} & \multicolumn{2}{c|}{0.7611} & 
    \multicolumn{2}{c|}{0.8033} & 0.7451 & 0.7600 & 0.5965 & 0.8623 & 0.6626 & 0.8079
    \\
    \multicolumn{1}{c|}{\multirow{4}{*}{}}&
    \multicolumn{2}{c|}{SVM-RBF}& \multicolumn{2}{c|}{0.8482} & \multicolumn{2}{c|}{0.8527} & 
    \multicolumn{2}{c|}{0.8992} & 0.9076 & 0.8218 & 0.6939 & $\underline{0.9523}^*$ & 0.7865 & \underline{0.8822}  \\
    \multicolumn{1}{c|}{\multirow{4}{*}{}}&
    \multicolumn{2}{c|}{SVM-TS} 
    & \multicolumn{2}{c|}{0.8216} & \multicolumn{2}{c|}{0.8241} & \multicolumn{2}{c|}{0.8829} & 0.8210 & 0.8220 & 0.7127 & 0.8951 & 0.7630 & 0.8570 \\
    \multicolumn{1}{c|}{\multirow{4}{*}{}}&
    \multicolumn{2}{c|}{PPC\_RNN+CNN}& \multicolumn{2}{c|}{0.6970} & \multicolumn{2}{c|}{0.7349} & 
    \multicolumn{2}{c|}{0.7701} & 0.7428 & 0.7091 & 0.4572 & 0.8931 & 0.5660 & 0.7905\\
    \cline{1-15}
    \multicolumn{1}{c|}{\multirow{5}{*}{M2}}&
    \multicolumn{2}{c|}{GLAN}& \multicolumn{2}{c|}{0.8546} & \multicolumn{2}{c|}{0.8612} & 
    \multicolumn{2}{c|}{0.9451} & 0.8501 & \underline{0.8831} & 0.8820 & 0.8328 & 0.8637 & 0.8536\\
    \multicolumn{1}{c|}{\multirow{5}{*}{}}&
    \multicolumn{2}{c|}{SureFact}& \multicolumn{2}{c|}{0.8629} & \multicolumn{2}{c|}{0.8631} & 
    \multicolumn{2}{c|}{0.9344} & 0.8397 & 0.8830 & \underline{0.8994} & 0.8396 & 0.8632 & 0.8547\\
    \multicolumn{1}{c|}{\multirow{5}{*}{}}&
    \multicolumn{2}{c|}{PSIN}& \multicolumn{2}{c|}{0.8601} & \multicolumn{2}{c|}{0.8577} & 
    \multicolumn{2}{c|}{0.9405} & 0.9240 & 0.7942 & 0.7933 & 0.9337 & 0.8526 & 0.8547\\
    \multicolumn{1}{c|}{\multirow{5}{*}{}}&
    \multicolumn{2}{c|}{UPFD-GCN}& \multicolumn{2}{c|}{0.8700} & \multicolumn{2}{c|}{\underline{0.8752}} & 
    \multicolumn{2}{c|}{0.9242} & $\underline{0.9286}^*$ & 0.8270 & 0.8180 & 0.9379 & \underline{0.8692} & 0.8782\\
    \multicolumn{1}{c|}{\multirow{5}{*}{}}&
    \multicolumn{2}{c|}{GCNFN}& \multicolumn{2}{c|}{\underline{0.8767}} & \multicolumn{2}{c|}{0.8729} & 
    \multicolumn{2}{c|}{\underline{0.9478}} & 0.8785 & 0.8621 & 0.8505 & 0.9076 & 0.8552 & 0.8803\\
    \cline{1-15}
    \multicolumn{1}{c|}{\multirow{3}{*}{Ours}}&
    \multicolumn{2}{c|}{DCE-RD ($\alpha=0$)}& \multicolumn{2}{c|}{0.8855} & \multicolumn{2}{c|}{0.8751} & 
    \multicolumn{2}{c|}{0.9493} & 0.9001 & 0.8412 & 0.8008 & 0.9323 & 0.8523 & 0.8864\\
    \multicolumn{1}{c|}{\multirow{3}{*}{}}&
    \multicolumn{2}{c|}{DCE-RD ($m=1$)}& \multicolumn{2}{c|}{0.8899} & \multicolumn{2}{c|}{0.9021} & 
    \multicolumn{2}{c|}{0.9579} & 0.9030 & 0.9042 & 0.9034 & 0.8990 & 0.9025 & 0.9007\\
    \multicolumn{1}{c|}{\multirow{3}{*}{}}&
    \multicolumn{2}{c|}{DCE-RD}& \multicolumn{2}{c|}{$\textbf{0.9031}^*$} & \multicolumn{2}{c|}{$\textbf{0.9122}^*$} & 
    \multicolumn{2}{c|}{$\textbf{0.9654}^*$} & 0.9140 & $\textbf{0.9103}^*$ & $\textbf{0.9111}^*$ & 0.9147 & $\textbf{0.9116}^*$ & $\textbf{0.9117}^*$\\
  \bottomrule
\end{tabular}
\end{table*}

\begin{table*}
  \caption{Comparison with baselines on the Ma-weibo dataset. M1 represents non-graph-based methods, and M2 refers to graph-based methods. Best baseline results are
highlighted with underlined text. Best overall results are marked by *. Statistically significant improvement (t-test over 5 different dataset splits, p-value$<$ 0.05) is highlighted with bold text. We evaluate the Accuracy (Acc.), F1 measure (F1), and AUC over the two categories and Precision (Prec.), Recall (Rec.), and F1 measure (F1) on each class.}

  \label{tab:MaWeibo Comparison}
  \begin{tabular}{c|cc|cccccccc|cc|cc}
    \toprule
    \multicolumn{1}{c|}{}&
    \multicolumn{2}{c|}{\multirow{2}{*}{Method}} &
    \multicolumn{2}{c|}{\multirow{2}{*}{ACC}} & \multicolumn{2}{c|}{\multirow{2}{*}{F1}} & \multicolumn{2}{c|}{\multirow{2}{*}{AUC}} & \multicolumn{2}{c|}{Prec} & \multicolumn{2}{c|}{Rec} & \multicolumn{2}{c}{F1}\\
    \cline{10-15}
    \multicolumn{1}{c|}{}&
    \multicolumn{2}{c|}{} & \multicolumn{2}{c|}{} & \multicolumn{2}{c|}{} & \multicolumn{2}{c|}{} & T & F & T & F & T & F\\
    \midrule
    \multicolumn{1}{c|}{\multirow{4}{*}{M1}}&
    \multicolumn{2}{c|}{DTC} & \multicolumn{2}{c|}{0.8136} & \multicolumn{2}{c|}{0.8132} & 
    \multicolumn{2}{c|}{0.8687} & 0.7751 & 0.8501 & 0.8306 & 0.7995 & 0.8019 & 0.8240
    \\
    \multicolumn{1}{c|}{\multirow{4}{*}{}}&
    \multicolumn{2}{c|}{SVM-RBF}& \multicolumn{2}{c|}{0.8861} & \multicolumn{2}{c|}{0.8861} & 
    \multicolumn{2}{c|}{0.9457} & 0.8710 & 0.8989 & 0.8795 & 0.8916 & 0.8752 & 0.8952  \\
     \multicolumn{1}{c|}{\multirow{4}{*}{}}&
    \multicolumn{2}{c|}{SVM-TS} 
    & \multicolumn{2}{c|}{0.8802} & \multicolumn{2}{c|}{0.8801} & \multicolumn{2}{c|}{0.9338} & 0.8576 & 0.9000 & 0.8827 & 0.8780 & 0.8700 & 0.8889 \\
     \multicolumn{1}{c|}{\multirow{4}{*}{}}&
    \multicolumn{2}{c|}{PPC\_RNN+CNN}& \multicolumn{2}{c|}{0.5237} & \multicolumn{2}{c|}{0.7090} & 
    \multicolumn{2}{c|}{0.6698} & 0.6177 & 0.5813 & 0.6654 & 0.7902 & 0.5399 & 0.7892\\
    \cline{1-15}
    \multicolumn{1}{c|}{\multirow{5}{*}{M2}}&
    \multicolumn{2}{c|}{GLAN}& \multicolumn{2}{c|}{\underline{0.9216}} & \multicolumn{2}{c|}{\underline{0.9245}} & 
    \multicolumn{2}{c|}{0.9648} & 0.9146 & 0.9325 & 0.9154 & $\underline{0.9282}^*$ & 0.9137 & \underline{0.9293}\\
    \multicolumn{1}{c|}{\multirow{5}{*}{}}&
    \multicolumn{2}{c|}{SureFact}& \multicolumn{2}{c|}{0.9167} & \multicolumn{2}{c|}{0.9162} & 
    \multicolumn{2}{c|}{\underline{0.9678}} & 0.8945 & $\underline{0.9366}^*$ & \underline{0.9392} & 0.8945 & \underline{0.9150} & 0.9140\\
    \multicolumn{1}{c|}{\multirow{5}{*}{}}&
    \multicolumn{2}{c|}{PSIN}& \multicolumn{2}{c|}{0.9038} & \multicolumn{2}{c|}{0.9076} & 
    \multicolumn{2}{c|}{0.9616} & 0.8492 & 0.9198 & 0.8632 & 0.9061 & 0.8528 & 0.9107\\
    \multicolumn{1}{c|}{\multirow{5}{*}{}}&
    \multicolumn{2}{c|}{UPFD-GCN}& \multicolumn{2}{c|}{0.9032} & \multicolumn{2}{c|}{0.9033} & 
    \multicolumn{2}{c|}{0.9655} & \underline{0.9208} & 0.8855 & 0.8837 & 0.9237 & 0.8981 & 0.8998\\
    \multicolumn{1}{c|}{\multirow{5}{*}{}}&
    \multicolumn{2}{c|}{GCNFN}& \multicolumn{2}{c|}{0.8790} & \multicolumn{2}{c|}{0.8803} & 
    \multicolumn{2}{c|}{0.9512} & 0.8921 & 0.8675 & 0.8711 & 0.8935 & 0.8773 & 0.8774\\
    \cline{1-15}
    \multicolumn{1}{c|}{\multirow{3}{*}{Ours}}&
    \multicolumn{2}{c|}{DCE-RD ($\alpha=0$)}& \multicolumn{2}{c|}{0.9167} & \multicolumn{2}{c|}{0.9182} & 
    \multicolumn{2}{c|}{0.9616} & 0.9073 & 0.9220 & 0.9351 & 0.9123 & 0.9180 & 0.9136\\
    \multicolumn{1}{c|}{\multirow{3}{*}{}}&
    \multicolumn{2}{c|}{DCE-RD ($m=1$)}& \multicolumn{2}{c|}{0.9274} & \multicolumn{2}{c|}{0.9238} & 
    \multicolumn{2}{c|}{0.9678} & 0.9356 & 0.9064 & 0.9261 & 0.9237 & 0.9215 & 0.9231\\
    \multicolumn{1}{c|}{\multirow{3}{*}{}}&
    \multicolumn{2}{c|}{DCE-RD}& \multicolumn{2}{c|}{$\textbf{0.9328}^*$} & \multicolumn{2}{c|}{$\textbf{0.9321}^*$} & 
    \multicolumn{2}{c|}{$\textbf{0.9812}^*$} & $\textbf{0.9319}^*$ & 0.9225 & $\textbf{0.9506}^*$ & 0.9201 & $\textbf{0.9292}^*$ & $\textbf{0.9309}^*$\\
  \bottomrule
\end{tabular}
\end{table*}

\subsection{Overall Performance}

Tables \ref{tab:McFake Comparison} and \ref{tab:MaWeibo Comparison} show the performance of DCE-RD and all the compared baselines on the MC-Fake and Weibo dataset, respectively.
From the results, we can make the following observations: 

\begin{itemize}
    
\item Deep learning methods perform significantly better than DTC and SVM-based models with hand-crafted features. It is reasonable as deep learning methods are capable to learn high-level representations of event graphs. 

\item GLAN, SureFact, PSIN, UPFD-GCN, GCNFN and our
method have better
performance than PPC\_RNN+CNN in all settings.
This is because PPC\_RNN+CNN only utilizes sequential information, while GLAN, SureFact, PSIN, UPFD-GCN, GCNFN and our method make use of propagation structure and base on graph neural networks. 

\item DCE-RD tends to have better performance than the previous methods. Table \ref{tab:McFake Comparison} and \ref{tab:MaWeibo Comparison} show that our method significantly outperforms all $9$ baselines (p-value $< 0.05$) in terms of six evaluation criteria on both benchmark datasets. For example, our method improves accuracy by $2.64\%$ on MC-Fake and $1.12\%$ on Ma-Weibo over the most competitive baselines. This shows that our proposal can effectively improve the interpretation and detection ability of the graph-based model.
Since DCE-RD not only effectively utilizes the textual content of posts and propagation structures but also focuses on discovering diverse counterfactual evidence by recognizing multiple counterfactual subgraphs in the rumor propagation graph. As diverse counterfactual evidence helps leverage different important substructures of rumor propagation to improve multi-view interpretation, which further improves the performance of graph-based rumor detection. 

\item Experimental results given in the row of DCE-RD ($m=1$) demonstrate that model with multiple counterfactuals outperforms the model with only one counterfactual. Relying on a single explanation is limiting if there exist other data that could be modified to achieve a more favorable outcome. Other results given in the row of DCE-RD ($\alpha=0$) demonstrate that the proposed diversity loss is critical for rumor detection as it could ensure the diversity of counterfactual evidence, which significantly improves the detection ability of DCE-RD by leveraging multi-view interpretation. 

\end{itemize}

\begin{figure}[htpb]
  \centering
    \subfigure[MC-Fake]{
  \begin{minipage}[c]{\columnwidth}
      \includegraphics[width=1\columnwidth]{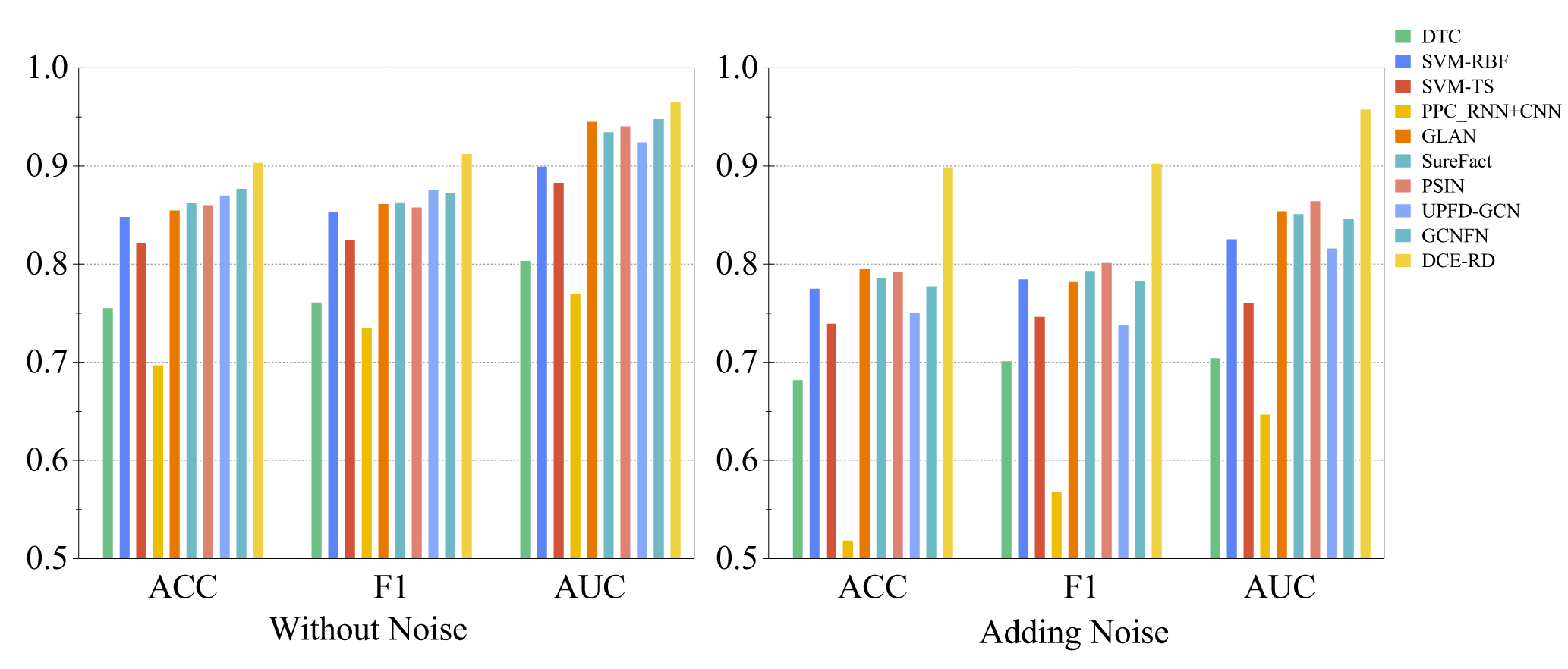}
  \end{minipage}}
  \subfigure[Ma-Weibo]{
  \begin{minipage}[c]{\columnwidth}
      \includegraphics[width=1\columnwidth]{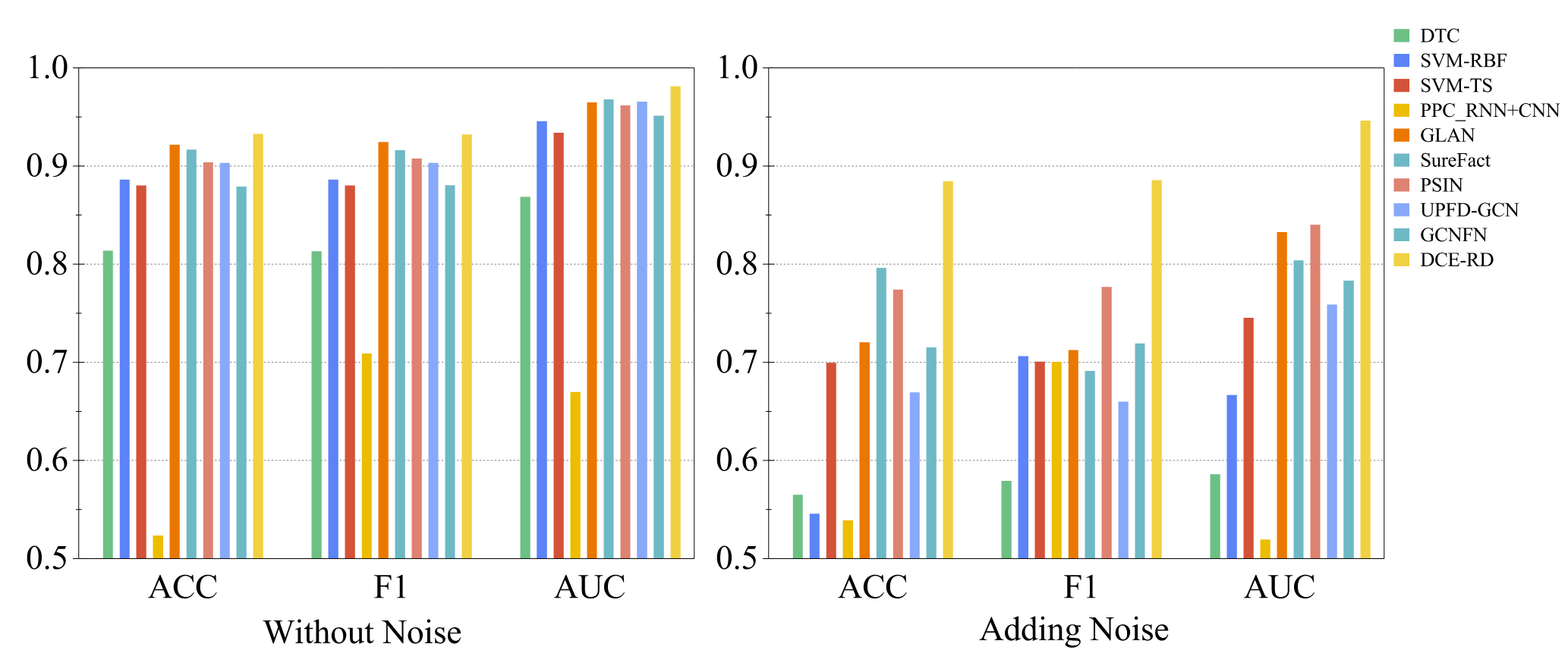}
  \end{minipage}}
  \caption{Overall performance of DCE-RD and baseline models without noise and adding Gaussian noise on two datasets.}
	\label{fig:Overall Performance}
	\vspace{0in}
\end{figure}

\subsection{Robustness}
Enhancing the robustness of the proposed method is important since GNNs usually suffer from random noise and adversarial attacks \cite{chang2020restricted,xie2023adversarially,gu2020implicit,chen2022understanding}.
Thus, we add Gaussian noise \cite{chen2015efficient} to the node features to investigate the robustness of our model under the noisy node features. Investigating the model robustness to noisy node features is important as many previous works \cite{zheng2016improving, yukawa2022stable, franceschirobustness} have done. Thus, we conduct the experiments under noisy node features to evaluate the robustness of our model. Gaussian noise refers to a type of noise whose probability density function obeys a Gaussian distribution (normal distribution) $\Phi \sim N(0,1)$. Specifically, we evaluate the robustness of all methods by quantifying the change in subgraphs after adding noise to the input graph. 
For an input graph $G$ and the set of sampled subgraphs $G_{sub}$, we generate a perturbed graph $G_p$ by adding random noise to the node features and randomly adding or removing some edges of the input graph such that the predictions for $G_p$ are consistent with $G$. Using the same method, we get the set of subgraphs $G_{sub}^p$ sampled on $G_p$. We then compute ACC, F1 and AUC on all compared baselines and our model DCE-RD, respectively. By observing the experimental results given in Figure \ref{fig:Overall Performance}, we find that the evaluation metrics of compared baselines dropped sharply after adding noise, but DCE-RD still presents a stable result. This indicates that DCE-RD is more robust and stable under noisy node features compared with the best baseline, which demonstrates that diverse counterfactual evidence not only improves the model performance but also guarantees the robustness of model.

\subsection{Early Rumor Detection}

\begin{figure}[htpb]
  \centering
  \subfigure[MC-Fake]{
  \begin{minipage}[c]{0.45\columnwidth}
      \includegraphics[width=\columnwidth]{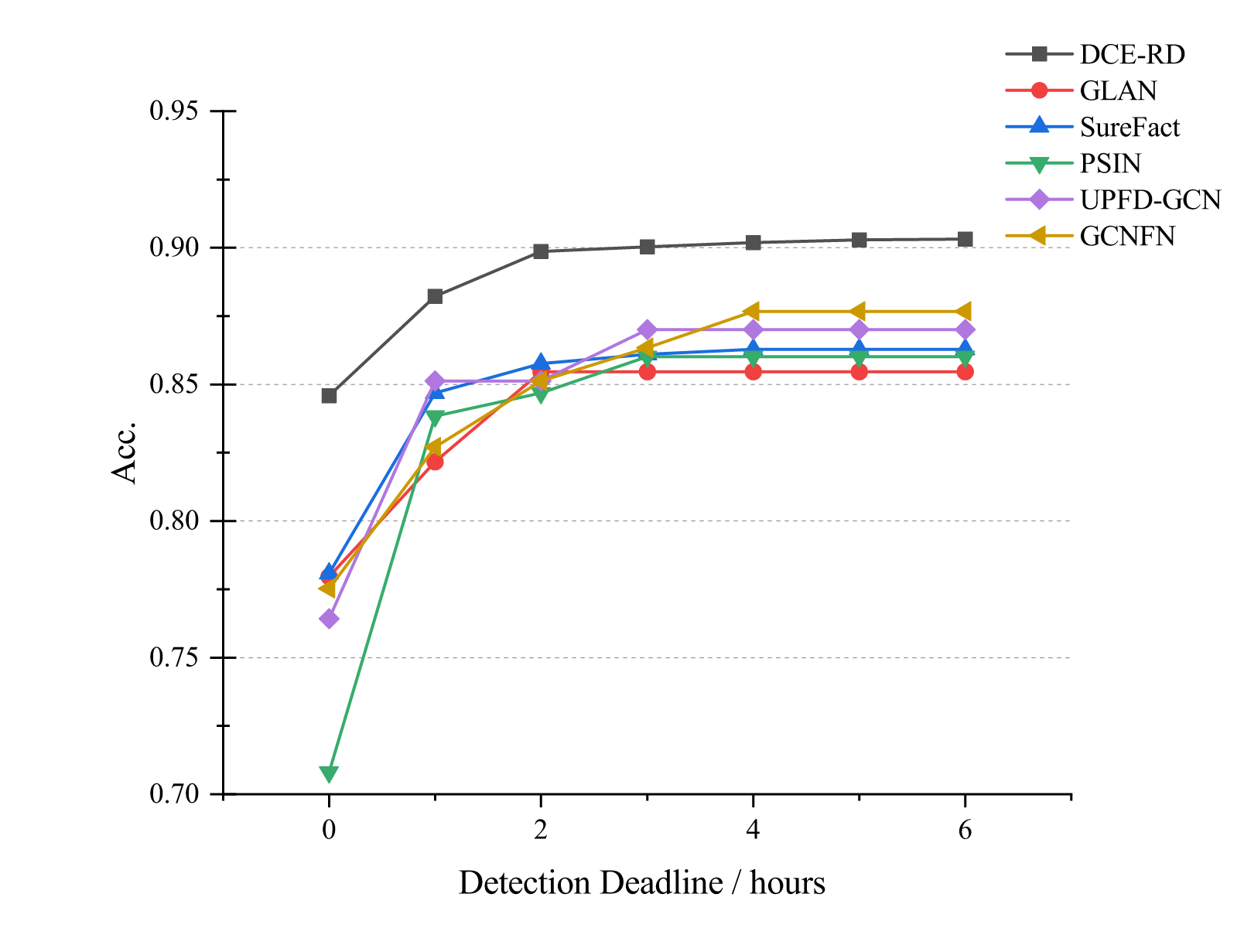}
  \end{minipage}}
  \subfigure[Ma-Weibo]{
  \begin{minipage}[c]{0.45\columnwidth}
      \includegraphics[width=\columnwidth]{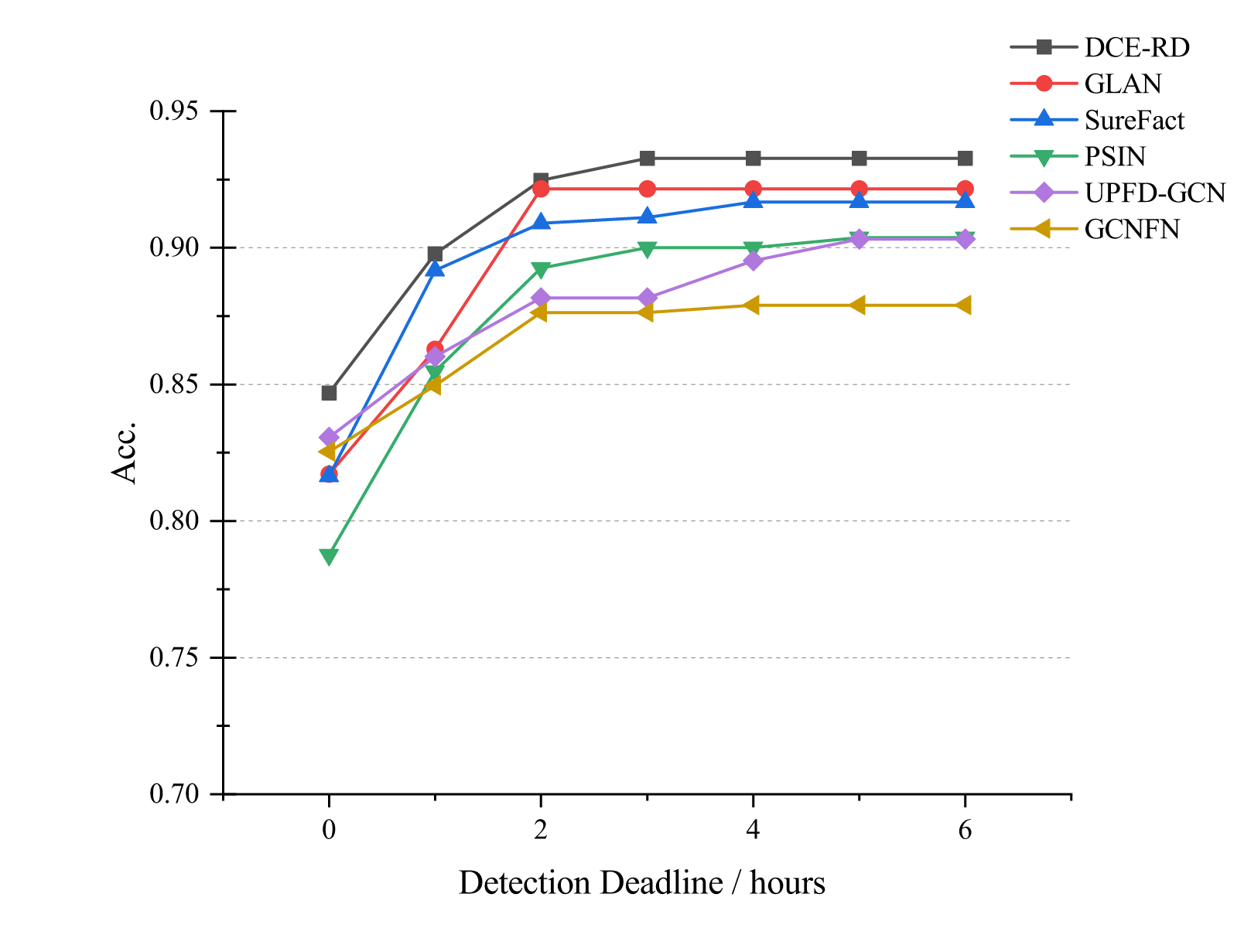}
  \end{minipage}}
  \caption{Results of rumor early detection on two datasets.}
	\label{fig:Early Detection Performance}
	\vspace{0in}
\end{figure}

Early rumor detection is another important metric to evaluate the quality of the model, which aims to detect rumors at the early stage of propagation \cite{liu2018early}. To establish an early detection task, we first set up a series of detection deadlines, such as $0,2,4,6$ hours. We then focus on evaluating the accuracy (ACC.) of DCE-RD and compared baselines only utilizing the posts released prior to the deadlines. 
Figure \ref{fig:Early Detection Performance} shows the performances of DCE-RD versus GLAN, SureFact, PSIN, UPFD-GCN, GCNFN at various deadlines on the datasets MC-Fake and Weibo. 
According to the results, it can be seen that DCE-RD achieves
relatively high accuracy at a very early period after the initial broadcast of source post. Besides, DCE-RD remarkably outperforms other models at every deadline, indicating that providing interpretable explanations and diverse counterfactual evidence is beneficial not only for long-term rumor detection but also for early detection.

\section{conclusion}

In this paper, we introduce DCE-RD, a diverse counterfactual evidence framework for rumor detection. We propose a subgraph generation strategy to discover different substructures of the event graph. Specifically, DCE-RD performs Top-K Nodes Sampling with Gumbel-max Trick on the event graph to generate a set of nodes that keeps prediction-relevant information. And our focus is on exploiting the diverse counterfactual evidence of an event graph to serve as multi-view interpretations,  which is generated by constraining the removal of these subgraphs to cause the change in rumor detection results. To further achieve interpretable and robust results, we formulate a diversity loss inspired by Determinantal Point Processes to ensure the diversity of counterfactuals, which are further aggregated for robust rumor detection results. 
Experiments including overall detection performance, robustness evaluation and early detection verify the effectiveness and superiority of DCE-RD compared with the state-of-the-art approaches.


\begin{acks}
This work was supported by the National Natural Science Foundation of China (NSFC) (Grant U2003111).
\end{acks}

\bibliographystyle{ACM-Reference-Format}
\balance
\bibliography{references}


\end{document}